\documentclass[sigplan]{acmart}

\usepackage{makecell}
\usepackage[most]{tcolorbox}
\usepackage{listings}
\usepackage{xcolor}

\setcopyright{none}
\renewcommand\footnotetextcopyrightpermission[1]{}

\author{Gelei Xu}
\affiliation{
  \institution{University of Notre Dame}
  \city{Notre Dame}
  \state{IN}
  \country{USA}
}
\email{gxu4@nd.edu}

\author{Ningzhi Tang}
\affiliation{
  \institution{University of Notre Dame}
  \city{Notre Dame}
  \state{IN}
  \country{USA}
}
\email{ntang@nd.edu}

\author{Xueyang Li}
\affiliation{
  \institution{University of Notre Dame}
  \city{Notre Dame}
  \state{IN}
  \country{USA}
}
\email{xli34@nd.edu}

\author{Toby Jia-Jun Li}
\affiliation{
  \institution{University of Notre Dame}
  \city{Notre Dame}
  \state{IN}
  \country{USA}
}
\email{toby.j.li@nd.edu}

\author{Zhi Zheng}
\affiliation{
  \institution{University of Notre Dame}
  \city{Notre Dame}
  \state{IN}
  \country{USA}
}
\email{zzheng3@nd.edu}

\author{Wei Jin}
\affiliation{
  \institution{Emory University}
  \city{Atlanta}
  \state{GA}
  \country{USA}
}
\email{wei.jin@emory.edu}

\author{Yiyu Shi}
\affiliation{
  \institution{University of Notre Dame}
  \city{Notre Dame}
  \state{IN}
  \country{USA}
}
\email{yshi4@nd.edu}

\begin{document}

\title{An Empirical Study of Agent Skills for Healthcare: Practice, Gaps, and Governance}

\begin{abstract}
Healthcare automation is shaped by local procedures and organizational constraints, so agent capabilities rarely transfer unchanged across settings. Agent skills, self-contained directories that package reusable procedures for AI agents, are emerging as a procedural layer for adapting healthcare agents across diverse healthcare settings. We present the first empirical analysis of healthcare agent skills, drawing on 557 healthcare-related skills filtered from 58,159 public skills on ClawHub and annotated along ten dimensions covering function, deployment context, autonomy, and safety. We find that public healthcare skills emphasize patient-facing workflow automation and monitoring rather than the diagnostic and treatment-oriented tasks foregrounded in healthcare-agent research; coverage of the healthcare lifecycle and specialized clinical inputs remains uneven; and general technical risk does not reliably capture clinical risk. These findings position healthcare skills as a procedural layer not yet addressed by current benchmarks and risk frameworks.
\end{abstract}


\maketitle

\section{Introduction}


Healthcare is a consequential domain for AI, but useful automation in healthcare is rarely defined by the task alone. The same nominal task can vary across hospitals, specialties, and patient populations because it is shaped by local procedures and organizational constraints. This variation creates a practical challenge for healthcare agents. Although agents can combine context, tools, and multi-step reasoning to support clinical decision-making and workflow automation~\cite{li2025cxr,liao2025reflectool,yu2025simulated}, their capabilities cannot be treated as fixed behaviors that transfer unchanged across settings. Therefore, AI agents for healthcare need a way to package procedures for reuse and adaptation without rebuilding the entire system.

Agent skills have recently emerged as one response to this packaging problem\footnote{https://platform.claude.com/docs/en/agents-and-tools/agent-skills/overview}. A skill is a self-contained directory with a \texttt{SKILL.md} file as its entry point. This file specifies the skill's name, description, and Markdown instructions, and may be bundled with scripts and reference files\footnote{https://agentskills.io/}. Agents discover skills from their metadata at startup and load the full instructions only when a task matches the skill description, a pattern the specification calls \emph{progressive disclosure}. This design represents agent behavior as a shareable, versioned, and inspectable artifact~\cite{wu2026agent}. Thus, skills provide a natural mechanism for encoding local healthcare procedures: they can lower the barrier for clinicians and domain experts to contribute while making agent behavior concrete enough for review and governance.

As skills become a procedural layer for healthcare agents, the public skill ecosystem offers a direct view of what developers actually package, rather than what the literature assumes should be. We use this view to characterize current healthcare skill \textbf{practice}, including how skills are authored, who they serve, and what tasks and inputs they are built around. Building on this, we examine the \textbf{gaps} revealed by the ecosystem, including its divergence from the task focus of healthcare-agent research and its uneven coverage of the healthcare lifecycle. We then turn to \textbf{governance} by asking how skills are distributed across autonomy levels and clinical impact, and whether they declare their own boundaries. To carry out this analysis, from a snapshot of 58{,}159 publicly listed skills on ClawHub\footnote{\url{https://clawhub.ai/}}, one of the largest public agent skill platforms, we identify 557 healthcare-related skills and annotate each along ten dimensions covering function, deployment context, autonomy, and safety (Table~\ref{tab:taxonomy}).

We highlight three findings. (1) Public healthcare skills are largely consumer-facing and concentrate on post-diagnosis workflow automation and research support, while healthcare-agent papers emphasize diagnosis- and treatment-oriented reasoning. (2) Coverage across the healthcare lifecycle is uneven: specialized clinical modalities, such as medical imaging and physiological signals, serve as primary inputs in fewer than 2\% of skills. (3) General technical risk does not reliably capture clinical risk: many skills with limited tool access still influence clinical judgment, and most lack explicit boundary statements. Together, these findings position healthcare skills as a procedural layer that current benchmarks and risk frameworks do not adequately capture.

\section{Corpus and Annotation}

\textbf{Corpus.} We construct our corpus from ClawdHub, accessed through its OpenClaw archive\footnote{\url{https://github.com/openclaw/skills}}. Starting from 58{,}159 publicly listed skills in an April 20, 2026 snapshot, we identify healthcare skills using a GPT-5-mini classifier applied to each skill's name and developer-authored description (full prompt in Appendix~\ref{appendix:filtering_prompt}). We define healthcare skills as those primarily concerned with clinical care delivery, medical operations, or life sciences in care contexts, including clinical documentation, diagnosis or treatment support, electronic health record workflows, medical coding, public health, and mental health care delivery. We exclude general fitness, wellness marketing, and lifestyle coaching skills without a clinical framing, defaulting to exclusion in ambiguous cases. After deduplication, the corpus contains 557 healthcare agent skills.

\textbf{Annotation.} We annotate each skill along ten dimensions covering its function, care-cycle stage, intended user, input modality, autonomy level, clinical impact, general technical risk, user vulnerability, and explicit safety-boundary statements. Following prior work that uses LLMs for structured annotation of code-related artifacts~\cite{tang2026programming}, we use GPT-5.4 as a scalable classifier over each skill's full \texttt{SKILL.md} file, including its name, description, and Markdown body. The taxonomy was developed through manual inspection of an initial sample, with category definitions refined to reduce ambiguity in healthcare-specific cases. The full taxonomy design is in Appendix~\ref{appendix:annotation_design}; Table~\ref{tab:taxonomy} lists all dimensions and their labels.


\section{Results}


\subsection{The Healthcare Skill Ecosystem: Lightweight, Concentrated, and Patient-Facing}

We characterize the healthcare skill ecosystem along five dimensions: artifact size, authorship, adoption, intended users, and linguistic or geographic framing.

\begin{figure}[htbp]
    \includegraphics[width=\columnwidth]{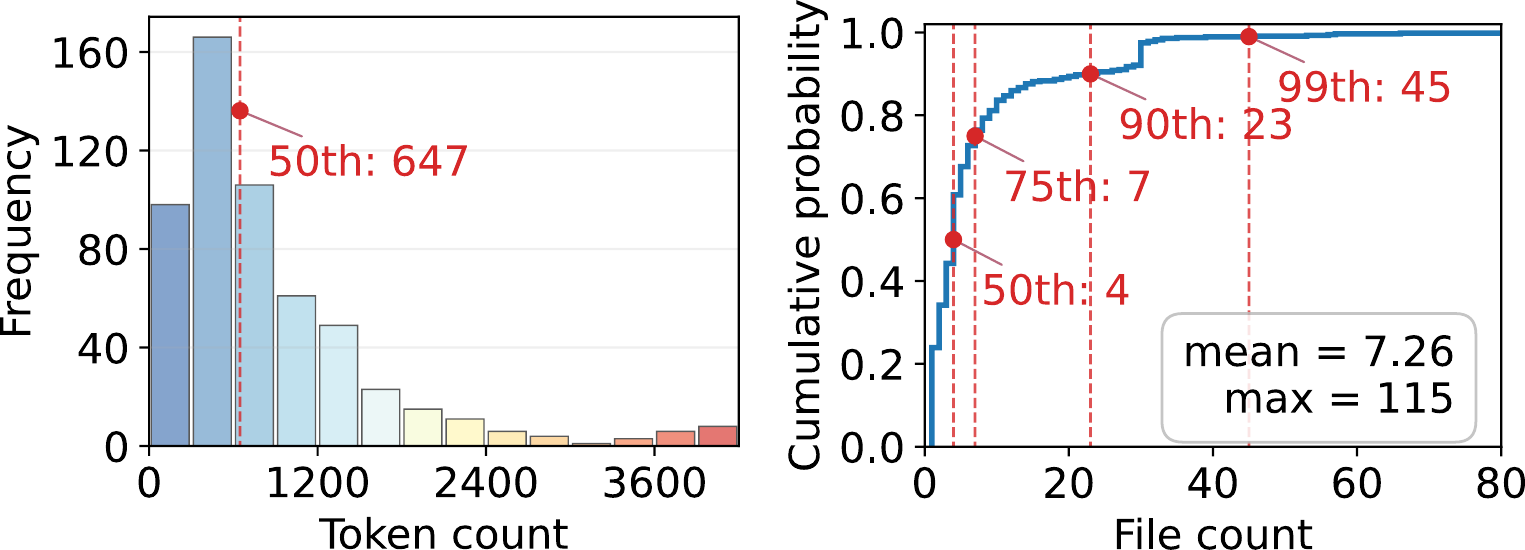}
    \caption{Distribution of healthcare skill size by token count (left) and file count (right).}
    \label{fig:token_file_count}
\end{figure}

\textbf{Most healthcare skills are lightweight procedural instructions, with only a small minority bundling multiple components into larger software artifacts.} The corpus has a median of 647 tokens and 4 files per skill (Figure~\ref{fig:token_file_count}), with long tails reaching 8,989 tokens and 115 files.

\textbf{Skill supply is disproportionately driven by a small group of highly active developers.} The 557 skills were authored by 233 unique contributors (mean: 2.39; median: 1). Authorship is highly skewed: the top-5 contributors account for 33\% of all skills, and the most prolific contributor alone authored 80 skills (14.4\%). We therefore interpret subsequent distributional results as patterns in the observed corpus, rather than evidence of broad developer demand.


\textbf{Healthcare skills show limited adoption relative to the broader skill ecosystem.} The three most-installed healthcare skills have 230, 195, and 187 installs, compared with 6,100, 4,100, and 4,100 for the top-3 skills platform-wide on OpenClaw. This 20--30$\times$ gap in install counts suggests an early-stage healthcare skill ecosystem: skills are being published, but few have attracted substantial user adoption.

\textbf{Public healthcare skills are predominantly patient-facing rather than clinician-facing.} Patients and general consumers form the largest intended-user category (37.1\%), surpassing any single professional category: researchers (22.4\%), clinicians (20.3\%), and hospital administrators (12.1\%). Caregivers and medical students together account for fewer than 8\%. This pattern contrasts with the clinical and institutional emphasis of much healthcare-agent research, suggesting that public skill development is currently driven more by consumer use cases than by professional workflows.

\textbf{The corpus is concentrated in English (65.0\%) and Chinese (25.7\%)}, with multilingual skills adding another 7.9\%; all other languages together account for fewer than 2\%. Separately, geographic signals are sparse: among the 209 skills with an identifiable target market, mainland China accounts for 128 and the United States for 42. The remaining 62.5\% of skills carry no clear geographic signal, suggesting that many are framed as general clinical or research procedures rather than market-specific implementations.

\subsection{Skills and Papers Emphasize Different Healthcare Work}
\label{sec:functional_distribution}

\begin{table}[htbp]
\centering
\small
\caption{Functional distribution of healthcare agent skills versus healthcare agent papers (taxonomy from \cite{xu2026comprehensive}). Mention rates exceed 100\% because each artifact may carry multiple function tags. $^{\dagger}$ marks functions present only in the paper taxonomy; --- indicates absence from the source taxonomy.}
\label{tab:functional_distribution}
\begin{tabular}{llrr}
\toprule
\textbf{Group} & \textbf{Function} & \textbf{Skills} & \textbf{Papers} \\
\midrule
\textit{Clinical} & Diagnosis & 22.4\% & \textbf{40.0\%} \\
 & Documentation & 19.0\% & 14.6\% \\
 & Treatment Planning & 15.1\% & 11.8\% \\
 & Consultation$^{\dagger}$ & — & \textbf{34.4\%} \\
 & Report Generation$^{\dagger}$ & — & 10.5\% \\
 & Triage$^{\dagger}$ & — & 5.9\% \\
\midrule
\textit{Administrative} & Workflow Automation & \textbf{33.2\%} & \textbf{37.2\%} \\
 & Health Commerce & 6.6\% & — \\
\midrule
\textit{Research} & Research Support & \textbf{28.9\%} & — \\
 & Simulation$^{\dagger}$ & — & 12.3\% \\
\midrule
\textit{Personal Health} & Health Education & \textbf{30.3\%} & 13.1\% \\
 & Patient Monitoring & 26.9\% & — \\
 & Mental Health & 9.0\% & — \\
\bottomrule
\end{tabular}
\end{table}

Table~\ref{tab:functional_distribution} shows the functional distribution of healthcare skills and healthcare agent papers~\cite{xu2026comprehensive}. Because the taxonomies differ, we interpret this comparison as a high-level contrast in emphasis rather than a category-by-category mapping.


\textbf{Skills favor procedural workflow automation over diagnostic reasoning.} Diagnosis is the leading category in the paper distribution, appearing in 40.0\% of papers but only 22.4\% of skills, likely reflecting established evaluation settings. In contrast, workflow automation is the most common skill category, appearing in 33.2\% of skills.

\textbf{Skills give greater visibility to user-facing and personal health use cases.}
Health education and patient monitoring appear in 30.3\% and 26.9\% of skills, respectively, and the skill taxonomy also includes mental health and health commerce. This pattern may partly reflect OpenClaw's role as a public skill ecosystem for personal agents, where developers can more readily package consumer-facing guidance, self-monitoring routines, and information support than procedures requiring integration with clinical systems.

\subsection{Healthcare Skills Concentrate in System Operations and Out-of-Clinic Patient Care}

\begin{figure}[htbp]
    \centering
    \includegraphics[width=\linewidth]{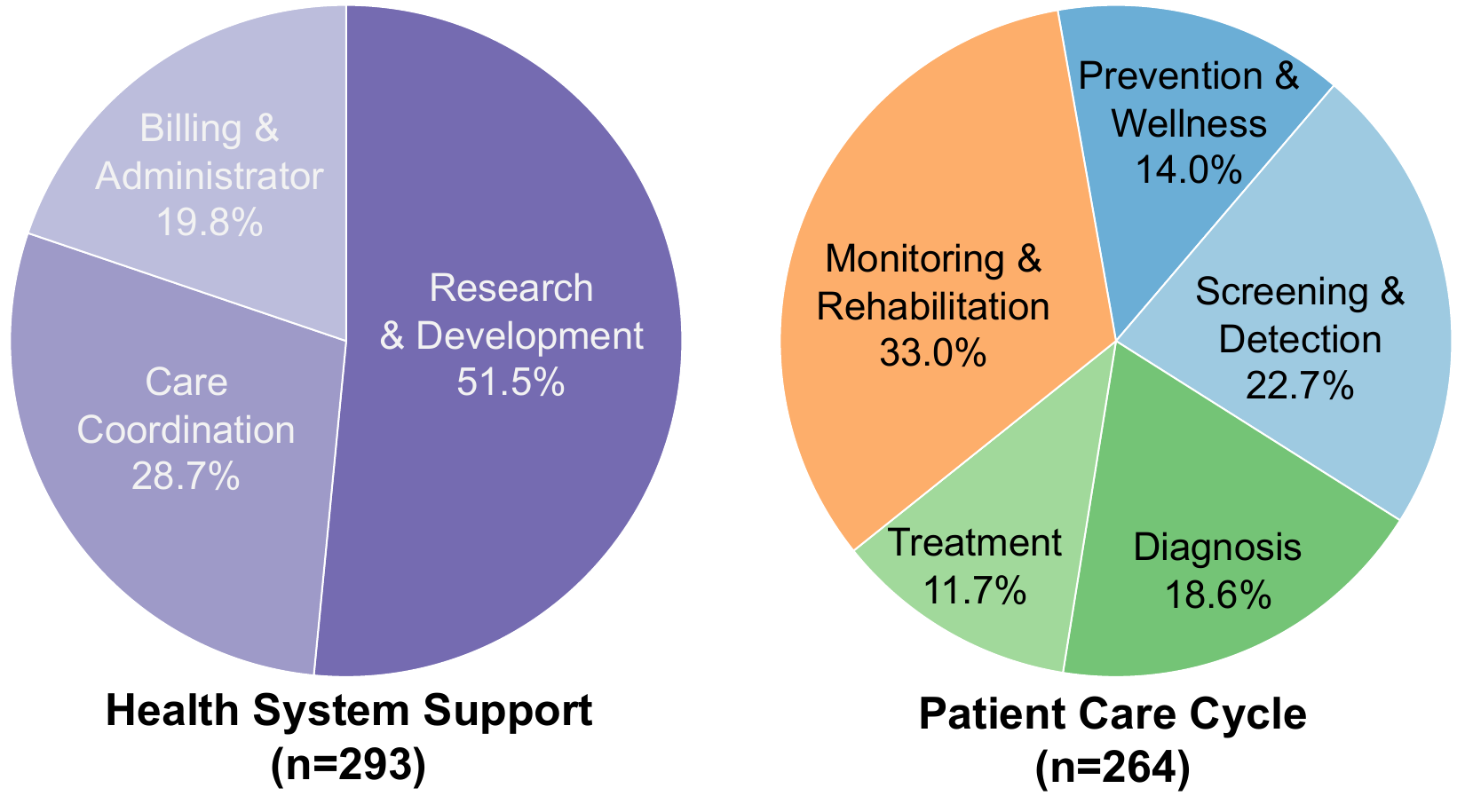}
    \caption{Distribution of skills across the healthcare lifecycle.}
    \label{fig:health_cycle}
\end{figure}

Whereas Section~\ref{sec:functional_distribution} describes what skills do, Figure~\ref{fig:health_cycle} describes where they sit in the healthcare lifecycle, splitting the corpus ($n=557$) into two families: patient care stages~\cite{devi2020narrative} ($n=264$) and health system operations ($n=293$).

\textbf{System-support functions account for more skills than patient-facing care stages (293 vs. 264), with Research \& Development the largest category} (151 skills, 27.1\% of the corpus). This concentration suggests that developers view research and coding tasks as especially suitable for procedural reuse, likely because they often rely on well-defined retrieval and synthesis routines with limited need for patient-specific context or real-time clinical judgment.

\textbf{Patient-facing skills cluster around screening and longitudinal support.}
Monitoring \& Rehabilitation is the largest patient-facing category (87 skills), followed by Screening \& Detection (60). These tasks may be more tractable for skill authors because they often occur outside acute clinical encounters, involve repeated structured interactions, and fit home or consumer settings. By contrast, diagnosis and treatment decisions require more patient-specific context and carry greater risks from error, which may discourage developers from encoding them as reusable procedures.

\subsection{Healthcare Skills Favor General-Purpose Inputs over Specialized Clinical Data}

\begin{figure}[htbp]
    \centering
    \includegraphics[width=\linewidth]{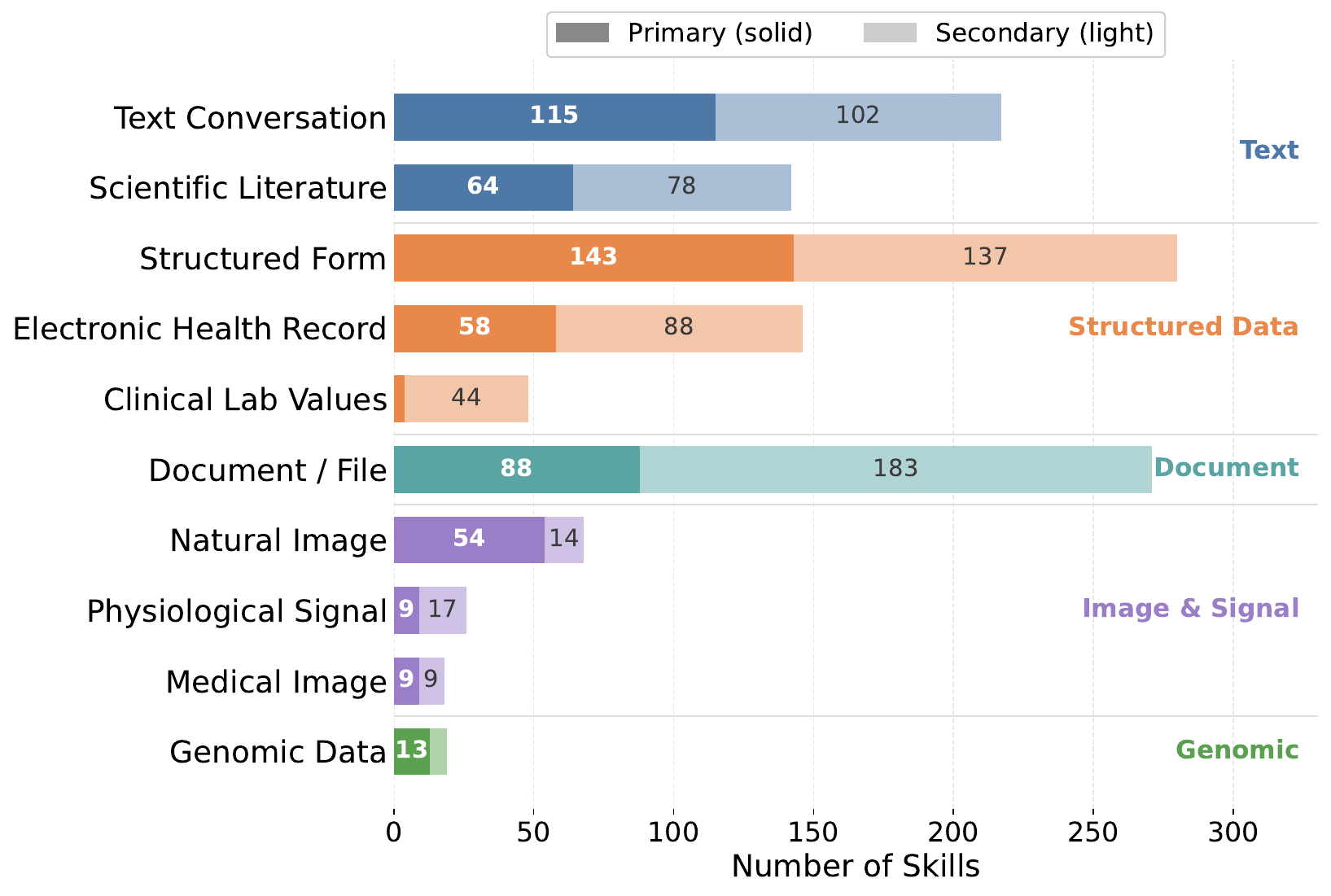}
    \caption{Input categories expected by healthcare skills.}
    \label{fig:data_modality}
\end{figure}

Figure~\ref{fig:data_modality} summarizes the input modalities expected by healthcare agent skills, distinguishing primary input channels from secondary supporting inputs. The distribution shows that current healthcare skills are built mainly around inputs that can be handled through standard LLM interfaces.

\textbf{Most skills rely on general-purpose input channels.}
Structured forms are the most common primary modality (143 skills), followed by text conversation (115), likely because forms support constrained intake and monitoring templates that standardize easily into reusable procedures.

\textbf{Documents are more often context than primary input.}
Document/file inputs appear as a secondary modality in 183 skills but as the primary modality in only 88. Skills often use uploaded reports or guidelines to contextualize workflows whose main interaction is driven by another channel.

\textbf{Specialized clinical modalities remain sparse.}
Medical images and physiological signals each appear as primary modalities in only 9 skills, while genomic data appears in 13 and clinical lab values in 44, roughly an order of magnitude fewer than forms, conversations, and documents. These modalities likely remain underrepresented because they require specialized infrastructure, dedicated models, or stronger validation than general LLM interfaces provide.

\subsection{Healthcare Skills Cluster at Delegated Execution with Mixed Clinical Stakes}
\begin{figure}[htbp]
    \centering
    \includegraphics[width=0.75\linewidth]{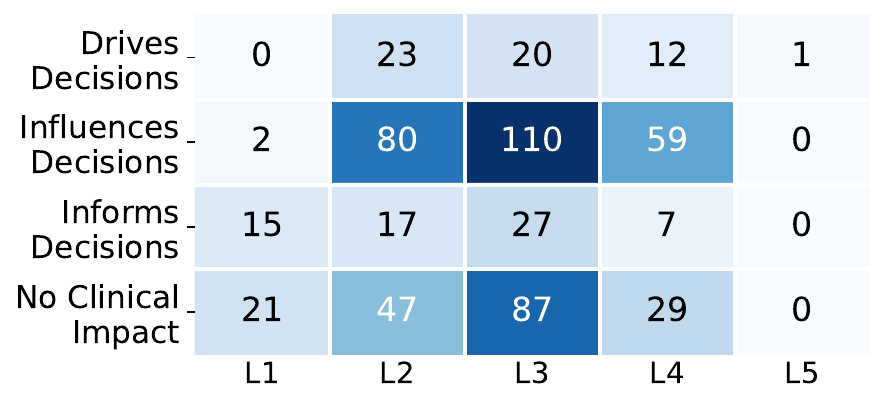}
    \caption{Joint distribution of clinical impact (rows, decreasing top-to-bottom) and autonomy level (columns, L1--L5 from lowest to highest) across all 557 healthcare skills. Each skill is assigned exactly one (impact, autonomy) pair.}
    \label{fig:autonomy_clinical_risk}
\end{figure}

Figure~\ref{fig:autonomy_clinical_risk} cross-tabulates each skill's autonomy level against its potential influence on clinical decision-making.

\textbf{L3 (delegated execution) dominates (244 skills), with the largest cell at Influences Decisions $\times$ L3 (110 skills).}
L3 is the modal autonomy column across every row. These L3 skills complete bounded digital tasks (e.g., producing summaries, processing structured inputs, or generating assessments) whose outputs may shape user judgment without real-world actuation, illustrating why autonomy alone is an incomplete proxy for healthcare risk.

\textbf{High autonomy is rare but concentrated in consequential cells.}
L4 contains 107 skills, and L5 contains only one. Most L4 skills fall under Influences Decisions (59) or No Clinical Impact (29), but 12 L4 skills and the sole L5 skill are classified as Drives Decisions. The L5 case is an emergency-guardian skill that assumes the user may be unable to respond and can execute preauthorized rescue actions, including recording, location broadcasting, contact notification, emergency calls, and multi-channel escalation. These L3 cases are small in number yet represent the clearest governance concern, combining high clinical impact with substantial delegated action.

\textbf{General technical risk is mostly moderate or safe.}
Under a general agent-risk taxonomy, 177 skills are classified as Safe, 269 as Moderate Risk, 103 as Privacy Risk, and 8 as Critical Risk. Critical-risk skills are uncommon, suggesting that most public healthcare skills do not request the most hazardous technical capabilities, such as destructive operations, financial actions, or arbitrary code execution. The distribution also shows the limit of general technical risk as a healthcare signal, since a technically safe or moderate skill can still produce outputs with clinical implications.

\textbf{Most skills lack explicit boundary statements.}
Only 163 of the 557 skills include an explicit disclaimer or scope statement. Reusable procedures may be invoked outside the context their authors intended, and without stated boundaries an agent has limited criteria for qualifying its output.
\section{Discussion}

The skill ecosystem we observe is partial, skewed, and clinically consequential in ways that current evaluation practices do not yet account for. We discuss four implications.

\textbf{Early skill development favors tasks with lower authoring and validation burden.}
Skills concentrate in administrative workflows, research support, and patient monitoring, while diagnosis and treatment appear less frequently. This pattern likely reflects asymmetric development costs across healthcare tasks. Diagnosis- and treatment-oriented skills require stronger clinical evidence, clearer liability boundaries, and validated patient-specific data. Workflow and research skills can often be authored with less sensitive data and evaluated through process-oriented outcomes, making them more accessible targets for early procedural reuse.

\textbf{Workflow-oriented skills reveal a gap between benchmarkable tasks and deployable procedures.}
Healthcare agent benchmarks often favor tasks with computable outcomes (e.g., diagnostic accuracy, medical question answering). Many public skills depend on local workflow fit and user role, and these procedures are harder to evaluate outside deployment contexts because their value depends on downstream action as well as output correctness.

\textbf{Skill governance should separate clinical impact from technical permission risk.}
General agent-risk frameworks often emphasize data access and state-changing tool use. Healthcare skills require a clinical-impact lens because text-only outputs can shape clinical judgment, especially for patient and consumer users. Our results show that many clinically consequential skills lack explicit boundary statements, and existing disclaimers often shift verification responsibility to users. Skill-based governance should attach review status, provenance, version history, and intended-use constraints to the skill artifact, while evaluating clinical impact alongside autonomy and tool access.

\textbf{Patient-facing skills require clearer quality signals and broader expert participation.}
Monitoring \& Rehabilitation is the largest patient-facing category in our corpus (87 skills). Chronic disease support and mental health skills are also present. These skills suggest a path for encoding personal health routines that depend on user context and recurring needs, although public availability does not indicate clinical appropriateness. Progress will require visible quality signals, such as validation status, intended-use boundaries, and expert review. It will also require lower barriers for clinicians and domain specialists to contribute procedural knowledge, since public skill ecosystems may otherwise remain shaped mainly by technically active developers.

\section{Conclusion}


This paper presents the first empirical analysis of 557 healthcare agent skills, showing that the public ecosystem develops unevenly across the healthcare lifecycle and that technical autonomy does not reliably capture clinical impact. Our corpus comes from a single platform at a single time point, and our annotations infer skill properties from developer-authored descriptions rather than deployment behavior. Future work should examine deployed skill performance and the robustness of these findings across patient populations. As agent skills become established artifacts in healthcare AI, treating them as concrete objects of review, rather than implicit model behavior, offers a practical surface for evaluation, governance, and domain expertise.

\bibliographystyle{ACM-Reference-Format}
\bibliography{reference}

\appendix

\section{Healthcare Filtering Prompt}
\label{appendix:filtering_prompt}

\begin{lstlisting}[basicstyle=\ttfamily\footnotesize, breaklines=true]
You are a strict classifier for agent skills.

You will receive: username, skill_name_folder, skill_name_md, and description.

Task: Decide whether this skill is primarily about HEALTHCARE (clinical, medical, health systems, life sciences used in care delivery) or clearly adjacent domains per the rules below.

Return ONE JSON object only with exactly these keys:
- "is_healthcare": boolean
- "reason": string, concise English citing what in the description supports your decision in one sentence.

Healthcare IN scope:
- Clinical care: diagnosis, treatment, SOAP notes, EHR/EMR, medical coding/billing, prior auth, clinical documentation, care pathways, triage, radiology workflow, lab orders/results interpretation support (when framed for clinical use)
- Providers & care delivery: hospitals, clinics, physicians, nurses, pharmacists, care teams, patient engagement for medical care
- Medical devices / regulated health software when the skill is about building or operating them in a care context
- Public health & epidemiology when clearly about population health surveillance, outbreak response, health policy implementation (not generic statistics unless tied to health)
- Mental / behavioral HEALTH care (therapy workflows, psychiatric care coordination) when clearly clinical or care-delivery oriented

OUT of scope:
- Pure SEO/marketing for clinics without clinical or operational healthcare substance
- Generic productivity, coding, finance, crypto, gaming, unless the description clearly centers on healthcare delivery

If the text is ambiguous, prefer "is_healthcare": false unless there is a clear healthcare delivery or clinical operations focus.
\end{lstlisting}

\section{Annotation Design}
\label{appendix:annotation_design}

\begin{table*}
\centering
\small
\caption{Annotation taxonomy. Each healthcare skill is annotated along eleven dimensions covering its function, deployment context, autonomy, risk, and metadata.}
\begin{tabular}{l l p{10cm}}
\toprule
\textbf{Dimension} & \textbf{Type} & \textbf{Values} \\
\midrule
Function & \makecell[tl]{Multi-label\\(1 primary, $\leq$2 secondary)} & clinical\_documentation, diagnosis\_support, treatment\_support, patient\_monitoring, mental\_health, administrative\_workflow, research\_support, drug\_discovery, health\_education, emergency\_response, health\_commerce \\
\addlinespace
Care cycle stage & \makecell[tl]{Multi-label\\(1 primary, $\leq$2 secondary)} & \textit{Patient-facing:} prevention\_wellness, screening\_detection, diagnosis, treatment, monitoring\_rehabilitation. \textit{System support:} care\_coordination, billing\_administration, research\_development \\
\addlinespace
Primary user & \makecell[tl]{Multi-label\\(1 primary, $\leq$2 secondary)} & patient\_consumer, clinician, researcher, hospital\_administrator, caregiver, medical\_student \\
\addlinespace
Input modality & \makecell[tl]{Multi-label\\(1 primary, $\leq$2 secondary)} & \textit{General:} natural\_language\_text, document\_file, general\_image, structured\_form. \textit{Specialist clinical:} medical\_imaging, ehr\_structured\_data, physiological\_signal, genomic\_data, clinical\_lab\_values \\
\addlinespace
Autonomy level & Ordinal scale (L1--L5) & passive\_tool (L1), active\_assistant (L2), delegated\_executor (L3), supervised\_actor (L4), autonomous\_agent (L5) \\
\addlinespace
Clinical decision impact & Ordinal scale & none, informs, influences, drives \\
\addlinespace
General risk & Ordinal scale (G0--G3) & safe (G0), privacy\_risk (G1), moderate\_risk (G2), critical\_risk (G3) \\
\addlinespace
Safety boundary & Structured & disclaimer\_present (boolean); disclaimer\_strength (strong / moderate / weak); disclaimer\_scope (multi-select from: general\_medical\_advice, diagnosis, treatment, emergency, professional\_referral) \\
\addlinespace
Language & Categorical & ISO 639-1 code or \textit{multilingual} \\
\addlinespace
Geography & Free text & Inferred from regulatory, system, or location signals in the skill description \\
\bottomrule
\end{tabular}
\label{tab:taxonomy}
\end{table*}

\textbf{Functional dimensions.} We assign each skill a \textit{primary function} category from eleven options, a \textit{primary care-cycle stage}, and a \textit{primary user role}. Care-cycle stages follow a two-tier structure: patient-facing stages ordered along the care pathway, and health-system support functions that operate across the continuum rather than at a single point. Each admits up to two secondary labels when a second function, stage, or role is substantively present.

\textbf{Behavioral dimensions.} \textit{Input modality} distinguishes general-purpose channels accessible to any LLM-based system from specialist clinical channels that require domain-specific pipelines. \textit{Autonomy level} is assessed on a single axis of user control, ranging from passive information retrieval to fully autonomous real-world action without user confirmation. A single-axis design ensures mutual exclusivity, addressing a limitation of taxonomies that conflate output type with execution modality. 

\textbf{Risk and safety dimensions.} Healthcare-specific risk is captured through \textit{clinical decision impact}, which measures whether a skill's output may inform, influence, or drive health-related decisions. We annotate this dimension separately from general technical risk and autonomy because clinical consequences may arise even when tool access or delegated action is limited. \textit{General risk} follows the framework of Liang~\emph{et al.}~\cite{ling2026agent}, classifying the potential for system-level harm through data access or state-changing operations. \textit{Safety boundary declaration} assesses whether the skill description contains an explicit disclaimer limiting the skill's clinical authority, characterized by its strength and scope.

\end{document}